\documentclass[10pt,twocolumn,letterpaper]{article}

\usepackage{wacv}
\usepackage{times}
\usepackage{epsfig}
\usepackage{graphicx}
\usepackage{amsmath}
\usepackage{amssymb}
\usepackage{color,soul}
\DeclareMathAlphabet\mathbfcal{OMS}{cmsy}{b}{n}
\usepackage{url}
\usepackage{multirow}
\usepackage{makecell}
\usepackage{makecell}
\usepackage{enumitem}
\usepackage{arydshln}

\wacvfinalcopy 

\def\wacvPaperID{706} 

\ifwacvfinal\fi
\begin{document}

\title{Unsupervised Domain Adaptation in Person re-ID via k-Reciprocal Clustering and Large-Scale Heterogeneous Environment Synthesis}

\author{Devinder Kumar$^{1}$ \hspace{1.cm} Parthipan Siva$^{2}$ \hspace{1.cm} Paul Marchwica$^{2}$ \hspace{1.cm} Alexander Wong$^{1}$\\
$^{1}$University of Waterloo \hspace{1.5cm} $^{2}$Sportlogiq\\
{\tt\small \{devinder.kumar,a28wong\}@uwaterloo.ca, \{parthipan,paul\}@sportlogiq.com}
}

\maketitle
\ifwacvfinal\thispagestyle{empty}\fi

\begin{abstract}
   An ongoing major challenge in computer vision is the task of person re-identification, where the goal is to match individuals across different, non-overlapping camera views. While recent success has been achieved via supervised learning using deep neural networks, such methods have limited widespread adoption due to the need for large-scale, customized data annotation.  As such, there has been a recent focus on unsupervised learning approaches to mitigate the data annotation issue; however, current approaches in literature have limited performance compared to supervised learning approaches as well as limited applicability for adoption in new environments.  In this paper, we address the aforementioned challenges faced in person re-identification for real-world, practical scenarios by introducing a novel, unsupervised domain adaptation approach for person re-identification. This is accomplished through the introduction of: i) k-reciprocal tracklet Clustering for Unsupervised Domain Adaptation (ktCUDA) (for pseudo-label generation on target domain), and ii) Synthesized Heterogeneous RE-id Domain (SHRED) composed of large-scale heterogeneous independent source environments (for improving robustness and adaptability to a wide diversity of target environments). Experimental results across four different image and video benchmark datasets show that the proposed ktCUDA and SHRED approach achieves an average improvement of  +5.7 mAP in re-identification performance when compared to existing state-of-the-art methods, as well as demonstrate better adaptability to different types of environments. 
\end{abstract}

\section{Introduction}

\begin{figure}[t]
\centering
\includegraphics[width=0.23\textwidth]{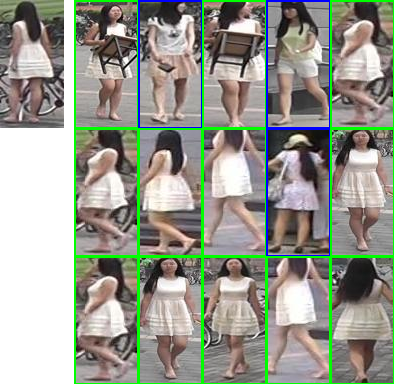}
\includegraphics[width=0.23\textwidth]{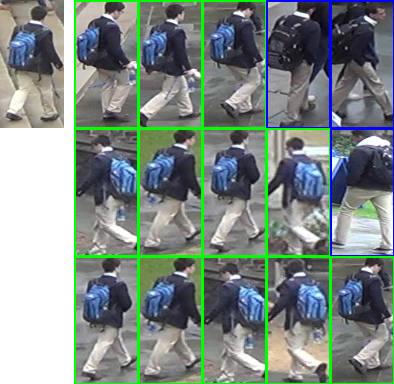}

\caption{Iterative adaptation to unlabelled target domain using the proposed ktCUDA approach. Result on test set after each iteration of adaptation on the unlabelled training set. Starting with direct knowledge transfer from the proposed SHRED source domain on the first row. Query on the left and the top-5 search result with green for correct match and blue for incorrect match. Image from Market-1501~\cite{zheng2015scalable} (left) and DukeMTMC-reID~\cite{gou2017dukemtmc4reid}  (right).
}
\label{fig:resultFig1}
\end{figure}

Person re-identification (re-ID) attempts to match an individual from one camera view across other, non-overlapping camera views \cite{Gong2014ReIDchallenge}. The most successful methods \cite{Yang2018local,Zeng2018hierarchical,kalayeh2018semantic} leverage deep learning via a supervised learning approach. Such supervised learning driven approaches assume the availability of a large, manually-labelled dataset of individuals across multiple cameras in the deployment environment (referred as the target domain). This assumption inherently limits the widespread adoption of person re-ID because of the cost and logistics needed for manually annotating data from the target domain, which is not practical in many real-world scenarios.

To overcome the reliance on a large, manually-labelled dataset from the target domain, two approaches have been proposed in recent literature: a pure unsupervised approach~\cite{li2018unsupervised}, and the more popular unsupervised domain adaptation approach~\cite{bak2018domain,zhu2017cycle,fan2018unsupervised,liu2017Stepwise,yu2017cross}. Both approaches rely on an unlabelled dataset from the target domain which is easily obtained by running tracking on the target domain. Furthermore, the unsupervised domain adaptation approach assumes the availability of a manually-labelled dataset from an independent source domain \cite{fan2018unsupervised,liu2017Stepwise,yu2017cross}, whereas the pure unsupervised approach does not require a manually-labelled source domain dataset. 

Without an independent source domain, the pure unsupervised approaches cannot function at all in a new target domain until they have learned the new environment. From a practical point of view, this is undesirable as the system is not able to function at all upon deployment. The unsupervised domain adaption methods on the other hand are pre-trained on an independent source domain and can function upon deployment by directly transferring models learned on the source domain (we refer to this as direct transfer). Starting from the direct transfer results, the system simply gets better as it adapts to the target domain (Fig.~\ref{fig:resultFig1}). The ability for immediate usage upon deployment makes such an unsupervised domain transfer approach very attractive from a practical point of view, but only if direct transfer performance is good and unsupervised domain adaptation can further improve the performance of the system. 


\begin{figure*}[t!]
\centering
\includegraphics[width=0.85\textwidth]{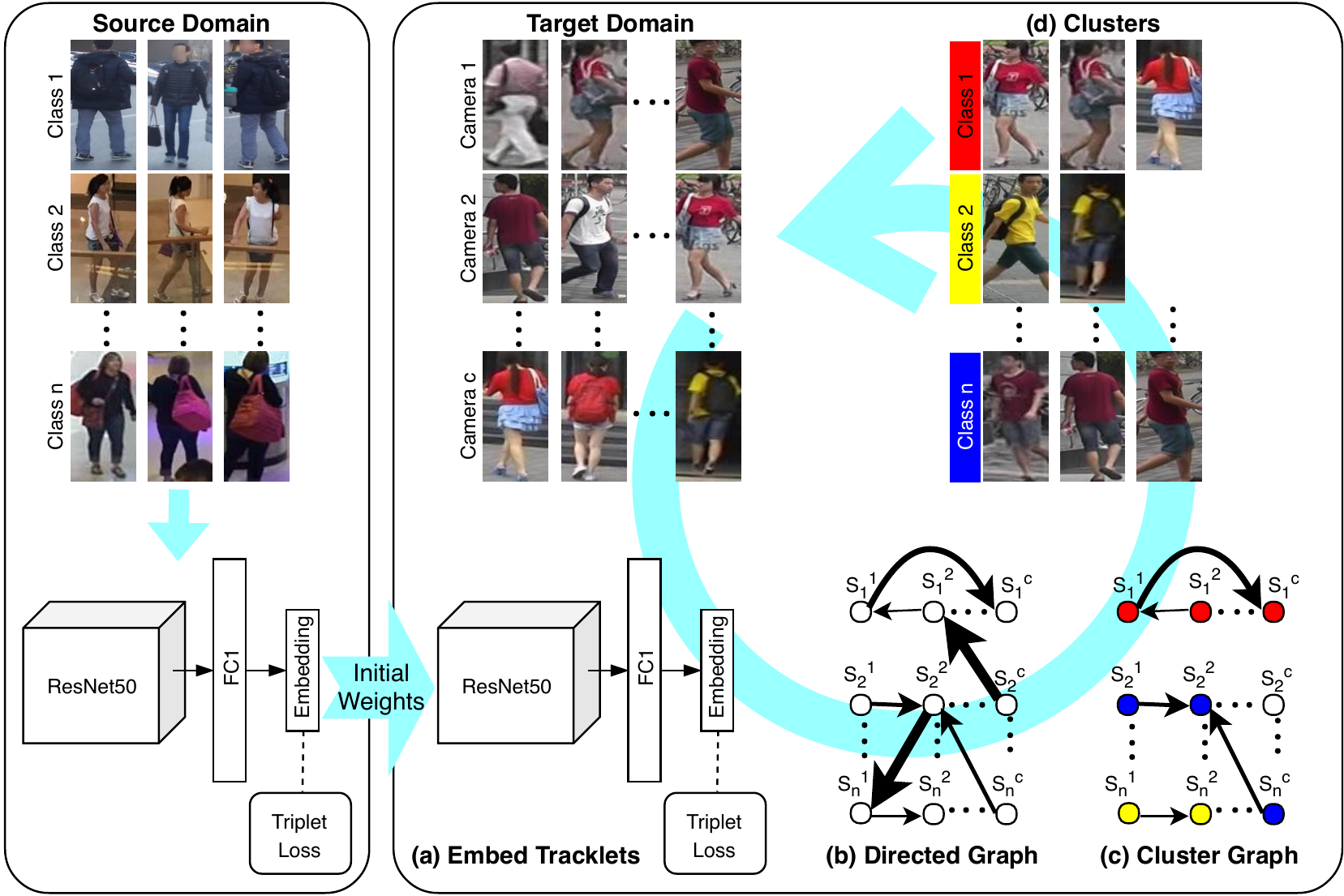}

\caption{Overview of the proposed k-reciprocal tracklet Clustering for Unsupervised Domain Adaptation (ktCUDA) in person re-ID. Given the proposed Synthesized Heterogeneous RE-id Domain (SHRED) as source domain, a ResNet-50 model with fully connected and embedding layers (DCNN) is trained with triplet loss. Once trained, the weights are used to initialize an iterative training on an unlabelled target domain. (a) DCNN is used to embed target domain tracklets to an embedding space. (b) Tracklet embeddings are used to form a directed graph, with each node representing a tracklet and each weighted connection representing how much the two tracklets belong to the same cluster.  (c) The directed graph is then thresholded based on weight to form clusters. (d) Tracklet images in each cluster formed in (c) are used to fine-tune the DCNN. This process (a)--(d) is repeated for $I$ iterations.}
\label{fig:system_diag}
\end{figure*}

There are two key limitations to existing unsupervised domain adaptation approaches~\cite{bak2018domain,zhu2017cycle,fan2018unsupervised,liu2017Stepwise,yu2017cross}. The first limitation is that the domain adaptation component of existing approaches either: i) only considers environmental style transfer between source and target domains \cite{bak2018domain,zhu2017cycle} and do not explicitly learn suitable features and distance metric for the target domain, or ii) directly transfer distance metric (typically Euclidean distance) \cite{fan2018unsupervised,liu2017Stepwise,yu2017cross} learned on the source domain to target domain for obtaining pseudo-labels on the target domain. Pseudo-labels are then used to learn suitable features and distance metric on the target domain. However, direct transfer of distance metric is not optimal due to differences in the source domain environment and target domain environment. 

The second limitation of existing unsupervised domain adaptation approaches is that they rely heavily on a limited real-world source domain. Typically, a single independent environment is used as a source domain \cite{li2018unsupervised,Wei2018GAN,song2018unsupervised,fu2018one,zhong2018generalizing} which doesn't capture enough variations in environments needed for domain adaptation. Some methods have attempted to augment the source domain with thousands of synthetic data with varying illuminations \cite{bak2018domain}, but other environmental variations outside of illumination are not captured. Finally, there are few works \cite{yu2017cross,bak2018domain,marchwica2018evaluation} that combine few different datasets in the source domain to obtain some variability. However, their performance before and after adaptation is generally noticeably lower as compared to the latest unsupervised person re-ID techniques~\cite{li2018unsupervised}.

In this work, we address the two aforementioned limitations of the current domain adaptation methods. First, we explore how to better leverage distance metrics that have been learned on the source domain to the target domain. Recently, k-reciprocal re-ranking~\cite{zhong2017reciprocal} has become a popular post-processing step for all supervised re-ID methods, where the k-reciprocal nearest neighbours are ranked higher than neighbours that minimize a distance metric and result in better performance. It was shown in~\cite{zhong2017reciprocal} to boost performance by $\sim10\%$ on the mean average precision (mAP). Motivated by the effectiveness of such an approach within the realm of supervised re-ID, we propose a k-reciprocal tracklet Clustering method for Unsupervised Domain Adaptation (ktCUDA), where k-reciprocal neighbours are used to assign pseudo-labels to the target domain.

Second, we investigate the construction of a source domain that captures large environmental variations i.e., large number of identities and environmental conditions to ensure the best results for direct transfer of source domain to target domain. To this end, we constructed the Synthesized Heterogeneous RE-id Domain (SHRED), the largest source domain used in domain adaptation person re-ID literature. We show that the proposed SHRED performs very well for the direct transfer scenario. When combined with the proposed ktCUDA, we show that state-of-the-art performance can be achieved for unsupervised domain transfer on several test datasets.

The main contributions of this paper are:
\begin{itemize}[noitemsep]
    \item \textbf{ktCUDA}, a novel k-reciprocal tracklet clustering algorithm for obtaining unsupervised pseudo-labels on the target domain.

    \item \textbf{SHRED}, a synthesized large-scale heterogeneous source domain that captures a wide set of environmental variations.

    \item A comprehensive analysis using both image and video datasets to show the performance of the proposed ktCUDA and SHRED, with full experimental results for direct transfer of knowledge from source domain to target domain as well as experimental results after domain adaptation.
\end{itemize}

\section{Related Works}

Unsupervised domain adaptation can take the form of environmental style (such as illumination) transfer between source and target domains~\cite{bak2018domain,zhu2017cycle,zhong2018generalizing} or iterative clustering and training based on distance metric transfer (typically Euclidean distance)~\cite{fan2018unsupervised,liu2017Stepwise,yu2017cross} between source and target domain. Our approach is an iterative clustering approach similar to \cite{fan2018unsupervised,liu2017Stepwise,yu2017cross}. However, unlike~\cite{fan2018unsupervised,yu2017cross}, which uses distance metric directly for clustering, we use k-reciprocal neighbours. The concept of using k-reciprocal neighbours in clustering for domain adaptation has been used in~\cite{liu2017Stepwise}. But in~\cite{liu2017Stepwise}, k-reciprocal neighbours are used to threshold potential cluster candidates then Euclidean distance is used during clustering. On the contrary, the proposed ktCUDA approach leverages k-reciprocal neighbour distance to perform spectral clustering without the reliance on distance metric during clustering.


\section{Methodology}
\label{SEC_method}

A common approach to unsupervised domain adaptation for person re-ID is to predict pseudo-labels for the unlabelled target domain using a deep convolutional neural network (DCNN) trained on the source domain and then fine-tune the DCNN for the target domain using the pseudo-labels~\cite{fan2018unsupervised,yu2017cross}. Typically, pseudo-labels are obtained by clustering~\cite{fan2018unsupervised,yu2017cross} using distance metrics on the samples in the target domain. Two problems with existing clustering based approaches~\cite{fan2018unsupervised, yu2017cross} are:
\begin{itemize}[noitemsep]
    \item The heavy reliance on distance metrics \cite{fan2018unsupervised,liu2017Stepwise} in the target domain using an embedding learned for the source domain. This results in poor clusters due to environmental differences between source and target domains making distance metrics unreliable between the domains.
    \item The use of a source domain with low environmental variability \cite{li2018unsupervised,Wei2018GAN,yu2017cross} or the reliance on synthetic environmental variability \cite{bak2018domain} result in a poor initial embedding for clustering.
\end{itemize}

We discuss our ktCUDA approach to overcome the strong reliance on distance metrics in Section~\ref{SEC_iterativeClustering}, and our SHRED approach to obtain the best source domain in Section~\ref{SEC_initalization}.

\subsection{Iterative Domain Adaptation}
\label{SEC_iterativeClustering}

Motivated to overcome the limitation of strong reliance on distance metrics in existing approaches ~\cite{fan2018unsupervised, yu2017cross}, we introduce a novel k-reciprocal tracklet Clustering for Unsupervised Domain Adaptation (ktCUDA). It has been shown in previous literature that leveraging k-reciprocal nearest neighbours to re-rank person re-ID search results, instead of the raw distances ranking, can result in a $\sim10\%$ boost in performance~\cite{zhong2017reciprocal,zheng2018pedestrian,li2018support}. Based on this observation, the proposed ktCUDA approach leverages the $k$ value in k-reciprocal nearest neighbours as the cost for joining tracklets into one cluster. This results in more accurate and robust clusters than using the raw distance between the two tracklets for the same reasons that re-ranking results in better person re-ID performance.


Our ktCUDA approach is illustrated in Fig.~\ref{fig:system_diag}. We iteratively fine-tune a DCNN on the unlabelled target domain by automatically obtaining labels using ktCUDA. More specifically, the following strategy was taken:

\begin{enumerate}[noitemsep]
    \item Transform the target domain tracklets to the embedding space using the DCNN (Fig.~\ref{fig:system_diag}(a)) \label{step_1}
    \item Cluster the tracklets using our k-reciprocal tracklet clustering approach (Fig.~\ref{fig:system_diag}(b-c)) \label{step_2}
    \item Use the clusters as the unsupervised labels for the tracklets and fine-tune our DCNN (Fig.~\ref{fig:system_diag}(d)) \label{step_3}
    \item Repeat steps \ref{step_1} to \ref{step_3} for $I$ iterations
\end{enumerate}

\subsubsection{k-Reciprocal Tracklet Clustering}
\label{SEC_clustering}

A tracklet is a short sequence of a tracked person in the video. Following findings of~\cite{zheng2016mars}, we represent the tracklet by the average embedding vector of the person bounding box on each frame of the tracklet. When we refer to a tracklet, we will be referring to the average embedding vector. The embedding is obtained by a DCNN; in our case, it is the same model as used in \cite{hermans2017defense} -- a ResNet-50 model with two additional fully connected layers as illustrated in Fig.~\ref{fig:system_diag}.

The unlabelled target domain
\begin{equation*}
    \mathbfcal{S} = \{\text{S}^1, \ldots, \text{S}^c, \ldots, \text{S}^N\}
\end{equation*}
\noindent is the set of all tracklet $\text{S}^c$ from $N$ cameras in the target domain and 
\begin{equation*}
    \text{S}^c = \{s_1^c, \ldots, s_t^c, \ldots, s_n^c\}
\end{equation*}
\noindent is the set of $n$ tracklet from camera $c$ and $s_t^c$ is the $t^{\text{th}}$ tracklet from camera $c$.

Given $\mathbfcal{S}$, the goal is to find clusters (i.e. subsets of $\mathbfcal{S}$) that represent a unique individual across multiple camera views. Two tracklets $s_t^c$ and $s_i^j$ with a small Euclidean distance $||s_t^c - s_i^j||$ will tend to be the same person if the DCNN was trained on the target domain using triplet loss. In this case, the DCNN was not trained on the target domain. 

A stronger argument is that if $s_t^c$ and $s_i^j$ are k-reciprocal neighbours of each other (for a small $k$ value) then the two tracklets will represent the same unique person \cite{qin2011hello}. Leveraging the idea of k-reciprocal nearest neighbours, we define a directed graph $\mathcal{G}$ where the weighted edges $\mathcal{E}$ represent k-reciprocal distance, the cost of assigning two tracklets to the same cluster. Clusters can then be formed on $\mathcal{G}$ to select tracklets representing a unique person.

\noindent\textbf{Graph Construction --} We define the $k_1-$nearest neighbours (i.e. the top-$k_1$ list) of $s_t^c$ as the closest tracklets in the target domain $\mathbfcal{S}$ excluding tracklets from camera $c$ (i.e. cross-camera closest tracklets using Euclidean distance):

\begin{equation}
    \text{top}(k_1,s_t^c) \in \mathbfcal{S} \setminus \text{S}^c
    \label{EQ_nn}
\end{equation}

Using \eqref{EQ_nn}, we construct a directed graph $\mathcal{G}=\{\mathcal{V},\mathcal{E}\}$ where vertices ($\mathcal{V}$) of the directed graph are representative of all the tracklets in the target domain: $s_t^c \in \mathbfcal{S}$ (i.e. $\mathcal{V}=\mathbfcal{S}$). Directed graph edges $e(s_t^c,s_i^j) \in \mathcal{E}$ are created from vertex $s_t^c$ to  all $s_i^j \in \text{top}(k_1,s_t^c)$. That is, we have $k_1$ directed edges starting from node $s_t^c$ to its $k_1-$nearest neighbours (Fig.~\ref{fig:system_diag}(a) illustrates a graph where $k_1=1$). Each edge $e(s_t^c,s_i^j)$ is given a weight, which we define as k-reciprocal distance:

\begin{equation}
    e(s_t^c,s_i^j) = k = \arg\min_k s_t^c \in top(k,s_i^j).
\end{equation}
\noindent In other words, we define $e(s_t^c,s_i^j)$, the distance between $s_t^c$ and $s_i^j$, as the minimum $k$ at which $s_t^c$ and $s_i^j$ are $k-$reciprocal neighbours of each other. For example if $s_i^j$ is the $1-$nearest neighbour of $s_t^c$ and $s_t^c$ is the $5-$nearest neighbour of $s_i^j$ then $e(s_t^c,s_i^j)=5$.

\noindent\textbf{Graph Clustering --} Given the graph $\mathcal{G}$ we form a new graph $\mathcal{G}'$ by cutting edge connections using threshold $K$:
\begin{equation}
    \label{EQ_graphWeightThresh}
    e(s_t^c,s_i^j) =
  \begin{cases}
    e(s_t^c,s_i^j) & \quad \text{if } e(s_t^c,s_i^j) \leq K\\
    \emptyset & \quad \text{if } e(s_t^c,s_i^j) > K
  \end{cases}
\end{equation}
\noindent where $e(s_t^c,s_i^j) = \emptyset$ means the connection between $s_t^c$ and $s_i^j$ has been removed (Fig.~\ref{fig:system_diag}(b) illustrates the graph $\mathcal{G}$ and Fig.~\ref{fig:system_diag}(c) illustrates the corresponding sparse graph $\mathcal{G}'$).

Due to these removal of connections, graph $\mathcal{G}'$ is a sparsely connected graph with a set of connected subgraphs $g' \subset \mathcal{G}'$. We define the cardinality of the connected subgraph $g'$ as:

\begin{equation}
    |g'| = \text{number of vertices in } g'
\end{equation}

From the sparse graph $\mathcal{G}'$ we create a valid cluster set $\mathcal{C}$ as the set of connected subgraphs with number of nodes (a.k.a tracklets) greater than $T$:

\begin{eqnarray}
    \label{EQ_graphCardinalityThresh}
    \mathcal{C} = \{g_{i=0,\ldots,m}'\} \quad \forall \quad &g_i' \in \mathcal{G}' \label{EQ_clusters} \\
    & |g_i'| > T \nonumber-0.4
\end{eqnarray}

\subsubsection{DCNN Fine-Tuning} 
\label{SEC_finetuning}
All images in the tracklets of a single cluster (i.e. subgraph $g'$) from the cluster set \eqref{EQ_clusters} are used as a unique class for the DCNN fine-tuning (Fig.~\ref{fig:system_diag}(d)). During fine-tuning, all the layers of DCNN are re-trained with unsupervised cluster data using batch hard triplet loss as per \cite{hermans2017defense}. For re-training, the weights from the previous iteration of domain adaptation are used as initialization.

\begin{table}[t]
	\caption{Composition of proposed SHRED source domain variants}
	\label{TAB:DatasetComposition}
	\begin{center}
		\setlength{\tabcolsep}{0.07cm}
		\scriptsize{
			\begin{tabular}{l||c|c|c|c|c|c}
				\hline
				Dataset & \makecell{SHRED \\1} & \makecell{SHRED \\2} & \makecell{SHRED \\3} & \# IDs & \# Images & \# Cameras \\
				\hline\hline
				
				3DPeS~\cite{baltieri2011_308} & $\checkmark$ & $\checkmark$ & $\checkmark$ & 164 & 951 & 8 \\ 
				[1pt]
				Airport~\cite{karanam2018airport} & $\checkmark$ & $\checkmark$ & $\checkmark$ & 1381 & 8660 & 6 \\
				[1pt]
				CUHK02~\cite{li2013locally} & $\checkmark$ & $\checkmark$ & $\checkmark$ & 1816 & 7264 & 10 \\
				[1pt]
				CUHK03~\cite{li2014deepreid} & $\checkmark$ & $\checkmark$ &  & 1467 & 14097 & 10 \\
				[1pt]
				DukeMTMC-reID~\cite{zheng2017unlabeled} & $\checkmark$ &  & $\checkmark$ & 1404 & 32948 & 8 \\
				[1pt]
				End-to-End~\cite{DBLP:journals/corr/XiaoLWLW16} & $\checkmark$ & $\checkmark$ & $\checkmark$ & 11934 & 34574 & N/A \\
				[1pt]
				GRID~\cite{GRIDDataset} & $\checkmark$ & $\checkmark$ & $\checkmark$ & 250 & 500 & 8 \\				
				[1pt]
				iLIDS-VID~\cite{Wang2016iLIDS} & $\checkmark$ & $\checkmark$ & $\checkmark$ & 300 & 42459 & 2 \\
				[1pt]
				MSMT17~\cite{Wei2018GAN} & $\checkmark$ & $\checkmark$ & $\checkmark$ & 3060 & 126142 & 15 \\
				[1pt]
				VIPeR~\cite{ViperDataset} & $\checkmark$ & $\checkmark$ & $\checkmark$ & 632 & 1264 & 2 \\
				\hline
				Market-1501*~\cite{zheng2015scalable} &  & $\checkmark$ & $\checkmark$ & 1501 & 32668 & 6 \\
				\hline
			\end{tabular}
			SHRED 1 is used to test on Market-1501, MARS and PRID datasets (22,408 IDs) \\
			SHRED 2 is used to test on DukeMTMC-reID dataset (22,505 IDs) \\
			SHRED 3 is used to test on CUHK03 dataset (22,442 IDS)
		}
	\end{center}
\end{table}

\subsection{Large-scale Heterogeneous Environment Synthesis}
\label{SEC_initalization}

While the iterative process described in Section~\ref{SEC_iterativeClustering} allows us to adapt a DCNN to a target domain, we still need initial DCNN weights to start with. Typically, an initial DCNN is trained on an independent source domain. The source domain can either be a single independent dataset \cite{li2018unsupervised,Wei2018GAN,song2018unsupervised,fu2018one,zhong2018generalizing}, a synthetic dataset \cite{bak2018domain}, or a combination of few independent datasets \cite{yu2017cross,bak2018domain,marchwica2018evaluation}.

An important consideration to keep in mind when considering adaptation from source domain to target domain is that we need embedding learned on the source domain to be as invariant as possible to environmental conditions such as lighting, background, etc. As such using a single independent environment \cite{li2018unsupervised,Wei2018GAN,song2018unsupervised,fu2018one,zhong2018generalizing} in the source domain is not ideal because network thus obtained will be too specific to the source domain. The use of synthetic source domain \cite{bak2018domain} can achieve invariance but only to the variables introduced in the generation of the synthetic data. Ideally, we would want the source domain to be created with data from many different actual environments as possible. With the nearly 30 different source domains that has been used since 2007 for person re-ID research~\cite{ridDataset}, it is possible to construct a source domain that has a wide variety of environmental variations.

Motivated to capture a wide of a set of environmental variations as possible, we construct a Synthesized Heterogeneous RE-id Domain (SHRED) from existing re-ID source domains under the following constraints:

\begin{itemize}[noitemsep]
    \item We avoid the use of source domains that have overlaps to ensure no one individual takes on two identities in the source domain. Some examples of source domains with overlap include CUHK02~\cite{li2013locally}-- CUHK01~\cite{li2012human}, DukeMTMC-reID~\cite{zheng2017unlabeled}-- DukeMTMC4ReID~\cite{gou2017dukemtmc4reid} and Market-1501~\cite{zheng2015scalable}-- MARS~\cite{zheng2016mars}.
    \item We avoid any gait domains such as~\cite{szheng_footprint_ICIP2011} in our SHRED source domain because they are staged in a studio environment with uniform background.
    \item We avoid source domains with less than or equal to 200 identities because they will be dwarfed by the larger source domains. Such small source domains include: Shinpuhkan~\cite{Shinpuhkan2014Dataset}, RAiD~\cite{das2014RaidDataset}, V47~\cite{wang2011v47}, HDA Person~\cite{figueira2014hdaDataset}, WARD~\cite{WARDDataset}, CAVIAR4ReID~\cite{Cheng:BMVC11}, MPR Drone~\cite{layne2014drone}, RPIfield~\cite{Zheng_2018_CVPR_Workshops}, PKU-Reid~\cite{MaLHWS16}, QMUL iLIDS~\cite{Zheng2009AssociatingGO}, SAIVT-SoftBio~\cite{quteprints53437}, ETH 1,2,3~\cite{conf/sibgrapi/SchwartzD09}.
    \item For each selected source domain, we combine data from training, validation and testing to ensure we have the largest possible variation in the source domain.
    \item For each source domain, we eliminate any individuals who doesn't appear in more than one camera as we want to ensure that the embedding learned from the source domain is for cross-camera comparison.
\end{itemize}

\noindent Based on the above constraints we are left with 12 source domains: 3DPeS~\cite{baltieri2011_308}, iLIDS-VID~\cite{Wang2016iLIDS}, VIPeR~\cite{ViperDataset}, PRID 2011~\cite{PRID2011Dataset}, GRID~\cite{GRIDDataset}, CUHK03~\cite{li2014deepreid}, Market-1501~\cite{zheng2015scalable}, DukeMTMC-reID~\cite{zheng2017unlabeled}, CUHK02~\cite{li2013locally}, MSMT17~\cite{Wei2018GAN}, Airport~\cite{karanam2018airport}, and End-to-End Deep Learning for Person Search~\cite{DBLP:journals/corr/XiaoLWLW16}. Of the 12 source domains, Market-1501 is a common source domain used for testing and it also overlaps with MARS video source domain which is another common source domain used for testing. As such, we leave Market-1501 out of the proposed SHRED to allow for testing on a large video and image datasets. Finally, PRID 2011 was excluded from our SHRED source domain because it only has 200 individuals appearing in multiple camera views.

The resulting SHRED \footnote{List of images in SHRED 1, 2, 3 as well as the DCNN weights trained on SHRED 1, 2, and 3 will be released.} source domain contains a heterogeneous mix of 10 different domains, with the details of the domain makeup shown in Table~\ref{TAB:DatasetComposition}. As stated in our selection constraints, the number of images used will be less than that originally reported for the respective domains because distractor identities or identities that don't appear in multiple cameras are removed in pre-processing.

Note that our proposed SHRED source domain contains DukeMTMC-reID and CUHK03. When we report results for DukeMTMC-reID, we remove it from our source domain and replace it with Market-1501 (i.e. SHRED 2 from Table~\ref{TAB:DatasetComposition}). Similarly, when reporting results for CUHK03, we remove it and replace it with Market-1501 (i.e. SHRED 3 from Table~\ref{TAB:DatasetComposition}).

\section{Experiment Setup}

\begin{table*}[t]
	\centering
	\setlength{\tabcolsep}{0.2cm}
	\caption{Direct transfer (SHRED) and unsupervised domain adaptation (SHRED+ktCUDA) performance on benchmark re-ID datasets compared to published methods. 
		$1^\text{st}$/$2^\text{nd}$/$3^\text{rd}$ best results are in \textbf{\color{red}red}/\textbf{\color{blue}blue}/\textbf{\color{cyan}cyan}. Multisource domain method in \color{magenta}magenta.}
	\label{tab:img_SOTA}
	\scriptsize{
	\begin{tabular}
		{l||c|c||c|c||c|c||c|c||c|c|c||c|c}
		\hline
		\multirow{2}{*}{Methods}
		& \multicolumn{2}{c||}{\makecell{Market \\-1501}\cite{zheng2015scalable}} 
		& \multicolumn{2}{c||}{MARS\cite{zheng2016mars}} 
		& \multicolumn{2}{c||}{CUHK03\cite{li2014deepreid}}                    	
		& \multicolumn{2}{c||}{\makecell{Duke\\ MTMC-reID}\cite{zheng2017unlabeled}}
		& \multicolumn{3}{c||}{PRID\cite{PRID2011Dataset}} & \multicolumn{2}{c}{Avg.}  \\ \cline{2-14}
			& R1	& mAP  & R1	& mAP	& R1	& mAP 	& R1	& mAP  & R1 & R5 & R20 & \textbf{R1} & \textbf{mAP}*
		\\ \hline \hline
		
		AML\cite{ye2007adaptive}		& 44.7  & 18.4	  & -       & -  & 31.4	 & -  & -     & -  & -  & - & -& - & -
		\\
		
		PTGAN \cite{Wei2018GAN}
		& 38.6       & - 
		& -       & - 
		& 24.8       & -                                  
		& 27.4       & -     
		& -       & -     & -& - & -
		\\ 
		PUL \cite{fan2018unsupervised}	& 44.7  & 20.1       & -      & - & -      & - & 30.4	& 16.4 & -  & - & -& - & -
		\\ 
		SPGAN+LMP \cite{deng2018image}          & 58.1  & 26.9   & -     & -   & -   & - & 46.4 & 26.2  & -  & -  & -   & - & -
		\\
        TJ-AIDL \cite{wang2018transferable}& 58.2  & 26.5   & -       & - & - 	& -        & 44.3 & 23.0 & -  & - & -& - & -
        \\
        
		HHL \cite{zhong2018generalizing}
		& 62.2    & 31.4  
		& -       & - 
		& -       & -                                  
		& \textbf{\color{cyan}46.9}	  & \textbf{\color{cyan}27.2}     
		& -       & -     & -& - & -
		\\ 
		TFusion \cite{zhong2018generalizing}
		& 60.8       & - 
		& -       & - 
		& -       & -                                  
		& -       & -    
		& -       & -     & -& - & -
		\\ 
		
		UnKISS \cite{khan2016unsupervised}	& - & -  & 22.3	   	& 10.6  & -  & -  & - & -    & 58.1 & 81.9  & 96.0  & - & -   \\ 
		SMP \cite{liu2017Stepwise}          & - & -  & 23.9   	& 10.5   & - - & - & - & -  & \textbf{\color{cyan}80.9} & \textbf{\color{cyan}95.6}  & \textbf{\color{blue}99.4}   & - & -  \\ 
		DGM+MLAPG \cite{ye2017dynamic}      & - & -  & 24.6     	& 11.8 	& -  & -  & - & - & 73.1  & 92.5 & \textbf{\color{cyan}99.0} 	& - & -	 \\ 
		DGM+IDE \cite{ye2017dynamic}        & - & -  & 36.8      & 21.3  & -  & -  & - & -   & 56.4 & 81.3  & 96.4   & - & -  \\
		RACE \cite{ye2018anchor}         	    & - & -  & 43.2      & 24.5 & -  & - & - & - & 50.6 & 79.4 & 91.8 & - & - \\ 
		DAL \cite{chen2018DAL}         	 & - & -  & \textbf{\color{cyan}46.8}      & 21.4  & -  & - & - & - & \textbf{\color{red}85.3}  &\textbf{\color{red} 97.0}    & \textbf{\color{red}99.6} & - & -\\ 
		TAUDL \cite{li2018unsupervised}
		& \textbf{\color{cyan}63.7}	& \textbf{\color{blue}41.2} 
		& 43.8	& \textbf{\color{cyan}29.1} 
		& \textbf{\color{red}44.7} 	& \textbf{\color{blue}31.2}  
		& \textbf{\color{red}61.7} 	& \textbf{\color{red}43.5}   
		& 49.4 & 78.7 & 98.9 & \textbf{52.7} & \textbf{36.3}
		\\
	    \color{magenta}JSTL\cite{xiao2016learning}		& 44.7  & 18.4  & -       & -  & 33.2   & -  & -      & -   & -  & - & -& - & -
		\\
		\color{magenta}CAMEL \cite{yu2017cross}		& 54.5  & 26.3   & -       & -  & \textbf{\color{cyan}39.4}	 & -  & - & - & -  & - & -& - & -
		\\
		\color{magenta}SyRI \cite{bak2018domain}
		& \textbf{\color{blue}65.7}       & -  
		& -       & - 
		& -       & -                                  
		& -       & -     
		& 43.0      & -     & -& - & -
		\\ 
		\hline
		\textbf{\color{magenta}SHRED}  
		& 53.9   & \textbf{\color{cyan}32.4}   
		& \textbf{\color{blue}53.3}   & \textbf{\color{blue}33.6}  
		& 28.5      & 26.1	 	
		& 40.9      & 24.5 
		& 76.4      & 94.4  &98.9 & \textbf{50.6} & \textbf{29.2}
		\\
		\textbf{\color{magenta}{SHRED+ktCUDA}}
				& \textbf{\color{red}68.6 }     & \textbf{\color{red}49.4}    
				& \textbf{\color{red}57.2}      &  \textbf{\color{red}36.0}  
		        & \textbf{\color{blue}44.4 }    & \textbf{\color{red}41.6}    
                & \textbf{\color{blue}58.7}		& \textbf{\color{blue}40.9} 
                & \textbf{\color{blue}84.3}  & \textbf{\color{blue}96.6}    &98.9 & \textbf{62.6} & \textbf{42.0}
		\\ \hline
		GCS \cite{chen2018gcs}({\em Sup.})	     & 93.5	& 81.6	      & - & - & 88.8	& 97.2   	& 84.9 	& 69.5 & -  & - & -  & - & -
		\\ 
		HDLF \cite{Zeng2018hierarchical}({\em Sup.})	     & -	& -	      & 86.4 & 79.3 & - & -   	& - 	& - & 95.7  & 99.1 & - & -  & - \\
		\hline
	\end{tabular}
	
	* PRID is excluded from average \textbf{mAP} because mAP is not a standard used to evaluate PRID~\cite{PRID2011Dataset}.
	}
\end{table*}

\begin{table*}[t]
	\centering
	\setlength{\tabcolsep}{0.2cm}
	\caption{Direct transfer (SHRED) and unsupervised domain adaptation (SHRED+ktCUDA) re-ranked ($rr$) \cite{zhong2017re} results. }
	\label{tab:rerank}
	\scriptsize{
	\begin{tabular}
		{l||c|c||c|c||c|c||c|c}
		\hline
		\multirow{2}{*}{Methods}
		& \multicolumn{2}{c||}{Market-1501\cite{zheng2015scalable}} 
		& \multicolumn{2}{c||}{MARS\cite{zheng2016mars}} 
		& \multicolumn{2}{c||}{CUHK03\cite{li2014deepreid}}                    	
		& \multicolumn{2}{c}{DukeMTMC-reID\cite{zheng2017unlabeled}}\\ \cline{2-9}
			& R1$_{rr}$  &mAP$_{rr}$   & R1$_{rr}$  &  mAP$_{rr}$ 	& R1$_{rr}$  &  mAP$_{rr}$  & R1$_{rr}$  &  mAP$_{rr}$ 
		\\ \hline \hline
		SHRED
			 & 57.4  & 43.7	& MSMT1753.4 & 41.1 & 37.6  &  37.7    & 47.0  & 38.0 \\ [0.5pt]
		SHRED + ktCUDA
		& 71.3	& 60.5	& 58.6 & 45.4 & 49.0  & 51.3    & 63.5      & 55.1 \\
		\hline
	\end{tabular}
	}
\end{table*}

The efficacy of leveraging the proposed ktCUDA and SHRED is investigated through a series of experiments across different image and video benchmark datasets.  The experimental setup in this paper is described below.

\noindent \textbf{Datasets --} We leverage three image datasets -- Market-1501, CUHK03 and DukeMTMC-reID -- to evaluate the proposed domain adaptation approach in a single-shot retrieval setting. In addition, we also use two video datasets -- MARS and PRID -- to evaluate the proposed approach in a multi-shot setting. When testing on Market-1501, MARS, and PRID, the source domain consists of SHRED 1 (Table~\ref{TAB:DatasetComposition}). When testing on CUHK03, the source domain consists of SHRED 3 (Table~\ref{TAB:DatasetComposition}). When testing on DukeMTMC-reID, the source domain consists of SHRED 2 (Table~\ref{TAB:DatasetComposition}).

In \cite{li2018unsupervised, liu2017Stepwise} for Market-1501, all images of an individual per camera are treated as a single tracklet and for MARS, a single tracklet per individual per camera is manually selected. For our experiments, we use the sequence ID in the Market-1501 dataset for tracklets, and for MARS we make no manual selection. As such, we have a harder and more realistic scenario of multiple tracklets of individuals per camera.

For all datasets, we follow the same test gallery-query split as in~\cite{li2018unsupervised}. All evaluation are done using the evaluation code provided with the datasets. For datasets without evaluation code, Market-1501 evaluation code was used.

\noindent \textbf{Implementation Detail --} Three parameters of our k-reciprocal tracklet Clustering for Unsupervised Domain Adaptation (ktCUDA) algorithm are: i) the number of domain adaptation iteration $I$, ii) the $k-$reciprocal distance threshold $K$ \eqref{EQ_graphWeightThresh}, and iii) the subgraph cardinality threshold $T$ \eqref{EQ_graphCardinalityThresh}. For all our experiments, we do at-least $I=2$ round of adaptation and only go above if the performance increases in the next round. We do early stopping only if number of cluster exceeds a soft upper bound on expected number unique individuals of 850. As the largest of the datasets contain around 700-750 identities, we number larger than that was chosen and hence 850 was picked. For deciding the values for $K$ and $T$, we chose the minimum number of cameras and cardinality that makes the domain iteration viable. Therefore, for datasets (Market-1501,MARS, and DukeMTMC-reID) with camera networks larger than two (that is an individual could potentially appear in more than 2 cameras), we set $K=2$ and $T=2$. For datasets (CUHK03 and PRID) with two camera network, we set $K=1$ and $T=1$ because we can't expect clusters larger than two since only two cameras exist in the network.

\noindent \textbf{Network Architecture --} All experiments were performed using the modified ResNet-50 network introduced in \cite{hermans2017defense}, which has an additional 1024 dimensional fully connected layer and a 128 dimensional embedding layer (see Fig.~\ref{fig:system_diag}).

\noindent \textbf{Training --} Training on the source domain is initialized with pre-trained ImageNet~\cite{deng2009imagenet} weights. Domain adaptation is initialized with weights trained on the source domain.

We keep the same training parameters provided by~\cite{hermans2017defense} with the exception of the number of iteration. We vary this based on our training data. For the source domain where we have much larger number of data due to the combination of several dataset, we set the number of iterations to $50,000$. For domain adaptation we use $25,000$ iteration for all datasets except PRID where we use $6,000$ iteration since it has far fewer images.

\begin{table}[t]
	\centering
	\setlength{\tabcolsep}{0.07cm}
	\caption{Comparison of SHRED direct transfer results with state-of-the-art unsupervised direct transfer methods on Market-1501. $1^\text{st}$/$2^\text{nd}$/$3^\text{rd}$ best results are in \textbf{\color{red}red}/\textbf{\color{blue}blue}/\textbf{\color{cyan}cyan}. Multisource domain method in \color{magenta}magenta.
	}
	\label{tab:source_domain}
	\scriptsize{
	\begin{tabular}
		{l||c|c|c}
		\hline
		{Methods}		& Source Domain & R1  & mAP	
		\\ \hline\hline
		TFusion\cite{lv2018Tfusion}	&   GRID &20.7 & -  \\
        TFusion\cite{lv2018Tfusion}	&   VIPeR &24.7	 & -   \\
		TFusion\cite{lv2018Tfusion}	  &   CUHK01 &29.4	 & -  \\
		PTGAN\cite{Wei2018GAN}	&   CUHK03 & 27.8  & -  \\
		HHL\cite{zhong2018generalizing}  &   CUHK03 &42.2		&20.3  \\
	    PTGAN\cite{Wei2018GAN}	&   DukeMTMC-reID & 33.5  & -  \\
		HHL\cite{zhong2018generalizing} 	&   DukeMTMC-reID            &44.6  	&20.6  \\
		T\&P\cite{song2018unsupervised}	&   DukeMTMC-reID & 46.8  &19.1  \\
		One-Shot\cite{Fu2018OneShot}	&   DukeMTMC-reID & 50.6	&\textbf{\color{cyan}23.7}  \\
        TJAIDL\cite{wang2018transferable}	&   DukeMTMC-reID & \textbf{\color{red}57.1}		& \textbf{\color{blue}26.2}  \\
        \hline
        \color{magenta}SyRI\cite{bak2018domain}	&   \makecell {CUHK03 + \\DukeMTMC-reID} & 44.7 & -  \\
        \color{magenta}SyRI\cite{bak2018domain}	&  \makecell{CUHK03 + \\ DukeMTMC-reID+SyRI} & \textbf{\color{blue}54.3}  & -  \\
        \hline
        \textbf{\color{magenta}ktCUDA}	&  SHRED & \textbf{\color{cyan}53.9}   & \textbf{\color{red}32.4}  \\
        \hline
	\end{tabular}
	}

\end{table}

\begin{table}[t]
	\centering
	\setlength{\tabcolsep}{0.07cm}
	\caption{Domain adaptation (ktCUDA) and direct transfer (SHRED) comparison for Market-1501. $1^\text{st}$/$2^\text{nd}$/$3^\text{rd}$ best results are in \textbf{\color{red}red}/\textbf{\color{blue}blue}/\textbf{\color{cyan}cyan}.Multisource domain method in \color{magenta}magenta.
	}
	\label{tab:source_domain_delta}
	\scriptsize{
		\begin{tabular}
		{l||c|c|c|c|c}
		\hline
       \multirow{2}{*}{Methods}
		& \multirow{2}{*}{\makecell{Source \\ Domain}}
		& \multicolumn{2}{c|}{Direct Transfer} 
		& \multicolumn{2}{c}{Domain Adapt.}                    	
        \\ \cline{3-6}
		& & R1  & mAP &  R1  & mAP
		\\ \hline\hline
		HHL\cite{zhong2018generalizing}   &CUHK03 &42.2 &20.3 &56.8 & 29.8 \\ 
		PTGAN\cite{Wei2018GAN}       &CUHK03 & -- & -- & 27.8 & -- \\ 
        \textbf{ktCUDA}	&CUHK03  &33.5 & 15.5
        &57.5 & \textbf{\color{cyan}35.2}  \\ \hline 
        PTGAN\cite{Wei2018GAN} 	&   DukeMTMC-reID  &33.5 &- &38.6 &- \\
        SPGAN+LMP\cite{deng2018image} 	&  DukeMTMC-reID  &43.1 &17.0 &58.1 &26.9 \\
		HHL\cite{zhong2018generalizing} 	&   DukeMTMC-reID  &44.6 &\textbf{\color{cyan}20.6} &62.2 &31.4 \\
		TJAIDL\cite{wang2018transferable}	& DukeMTMC-reID &\textbf{\color{red}57.1} &\textbf{\color{blue}26.2} &58.2 &26.5  \\

		\textbf{ktCUDA}  &  DukeMTMC-reID & 40.3 &17.6 &56.0 & 32.6  \\
        \hdashline
        TAUDL\cite{li2018unsupervised}*	&   None &- &- &\textbf{\color{cyan}63.7} & \textbf{\color{blue}41.2} \\
        \hline
        \color{magenta}CAMEL 	&   7set* & 41.4  & 14.1 & 54.5 & 26.3  \\ 
        \color{magenta}SyRI 	&   \makecell{CUHK03+\\DukeMTMC-reID+SyRI} & \textbf{\color{cyan}44.7} & -- & \textbf{\color{blue}65.7} & -- \\ 
        \textbf{\color{magenta}ktCUDA}	&  SHRED  & \textbf{\color{blue}53.9}  &\textbf{\color{red}32.4} & \textbf{\color{red}68.6}  &\textbf{\color{red}49.4}  \\
        \hline
	\end{tabular}
	
	7set*: VIPeR, CUHK01, CUHK03, PRID, 3DPeS, i-LIDS and Shinpuhkan.
	}
\end{table}

\section{Results and Discussion}

To compare the proposed ktCUDA and SHRED, we use the common Cumulative Matching Characteristic (CMC) and mean Average Precision (mAP) metrics. We evaluate against the state-of-the-art methods for domain adaption (where unlabelled target domain is used for training) and direct transfer (where target domain data is not used at all).

\subsection{Domain Adaptation}

The result of the proposed k-reciprocal tracklet Clustering for Unsupervised Domain Adaptation (ktCUDA) algorithm can be found in Table~\ref{tab:img_SOTA} (indicated as \textbf{SHRED + ktCUDA}) with comparison to existing state-of-the-art approaches.
It can be clearly observed that the proposed ktCUDA approach is the state-of-the art method for Market-1501 (\textbf{+8.2} mAP), MARS (\textbf{+6.9} mAP) and CUHK03 (\textbf{+10.4} mAP) datasets based on mAP amongst the tested methods. We also get competitive performance to state-of-the-art methods on DukeMTMC-reID and PRID datasets. 

In Table~\ref{tab:img_SOTA}, we present the average rank-1 (\textbf{R1}) and mean average precision (\textbf{mAP}) across all five test datasets as summary metrics. Based on the average performance, ktCUDA is \textbf{+9.9 R1} and \textbf{+5.7 mAP} better the current state-of-the-art. Finally, the efficacy of ktCUDA is shown by the observation that it is the only method that is consistently ranked as the best or competitive second best method on all five test datasets.

For the sake of completeness, we also present the re-ranked \cite{zhong2017re} results in Table~\ref{tab:rerank}.

\noindent \textbf{Comparison to multi-source domain methods --} While the general performance of SHRED+ktCUDA is consistently in the top two across all test sets it is worth looking at its performance relative to other multi-source domain methods (highlighted in magenta in Table~\ref{tab:img_SOTA}). Comparing to CAMEL, SHRED+ktCUDA uses 10 datasets versus CAMEL which uses 7 datasets and SHRED+ktCUDA outperforms CAMEL method. However, SHRED+ktCUDA has $\sim 250k$ images in the source domain compared to CAMEL's $\sim 45k$ images. Comparing to SyRI, which uses more than 1.6 million synthetic images and $\sim 45k$ real world images, SHRED+ktCUDA still out performs SyRI, thus motivating the need for real-world diverse images over synthetic images.



\subsection{Direct Transfer (SHRED without ktCUDA)}

It can be observed that the proposed SHRED source domain is quite effective across all test datasets as seen in Table~\ref{tab:img_SOTA} (indicated as \textbf{SHRED}). In particular, the performance on MARS dataset stands out. For MARS, our direct transfer results are \textbf{+4.5} mAP better than state-of-the-art domain transfer methods, even when these methods use unlabelled MARS data in the training.

A comparison of the proposed SHRED source domain for direct transfer with domain transfer methods on the Market-1501 dataset based on previously published results in literature can be found in Table~\ref{tab:source_domain}. As expected, we can see the proposed SHRED source domain outperforms existing source domains on mAP by a large margin.

Considering synthetic dataset augmentation (SyRI \cite{bak2018domain}) results in Table~\ref{tab:source_domain}, we observe that its Rank-1 result is slightly higher than the proposed SHRED source domain. Unfortunately this analysis is not conclusive without mAP. However, \cite{bak2018domain} also report Rank-1 result for single-shot re-ID on PRID dataset as 15\%. Our direct transfer for PRID single-shot re-ID gets a rank-1 accuracy of 22\%. Therefore, while synthetic data augmentation is good for giving some variability, real data from multiple sources is ultimately better.

\subsection{Domain adaptation boost}

For the three best source domain direct transfer results in Table~\ref{tab:source_domain}, TJ-AIDL \cite{wang2018transferable}, SyRI \cite{bak2018domain} and the proposed ktCUDA, we look at the improvement achieved by domain transfer over the direct transfer results in Table~\ref{tab:source_domain_delta}. From Table~\ref{tab:source_domain_delta} we observe, of the methods with best direct transfer results, the proposed method has best domain adaptation boost for mAP. Of particular importance is the SyRI method which uses a much larger source domain than SHRED+ktCUDA and has similar direct transfer accuracy as SHRED+ktCUDA. From the same starting point, SHRED+ktCUDA was able to achieve higher Rank-1 result than SyRI showing that ktCUDA is very effective strategy for domain adaption.

This shows that while a heterogeneous source domain is very effective at giving a good initialization, the proposed ktCUDA is also well-suited for adapting to a new domain.

We test our proposed ktCUDA approach with DukeMTMC-reID as the source domain and Market-1501 as the target domain as well in Table~\ref{tab:source_domain_delta}. This tests how well ktCUDA works for domain adaptation without using our proposed SHRED as the source domain. We can see that the proposed ktCUDA approach outperforms existing domain adaptation methods that use DukeMTMC-reID as the source domain. Furthermore, when combined with SHRED the proposed ktCUDA approach can get a significant boost over existing state-of-the-art methods.

\subsection{k-Reciprocal Tracklet Clusters}

\begin{figure}[t]
\centering
\includegraphics[width=0.45\textwidth]{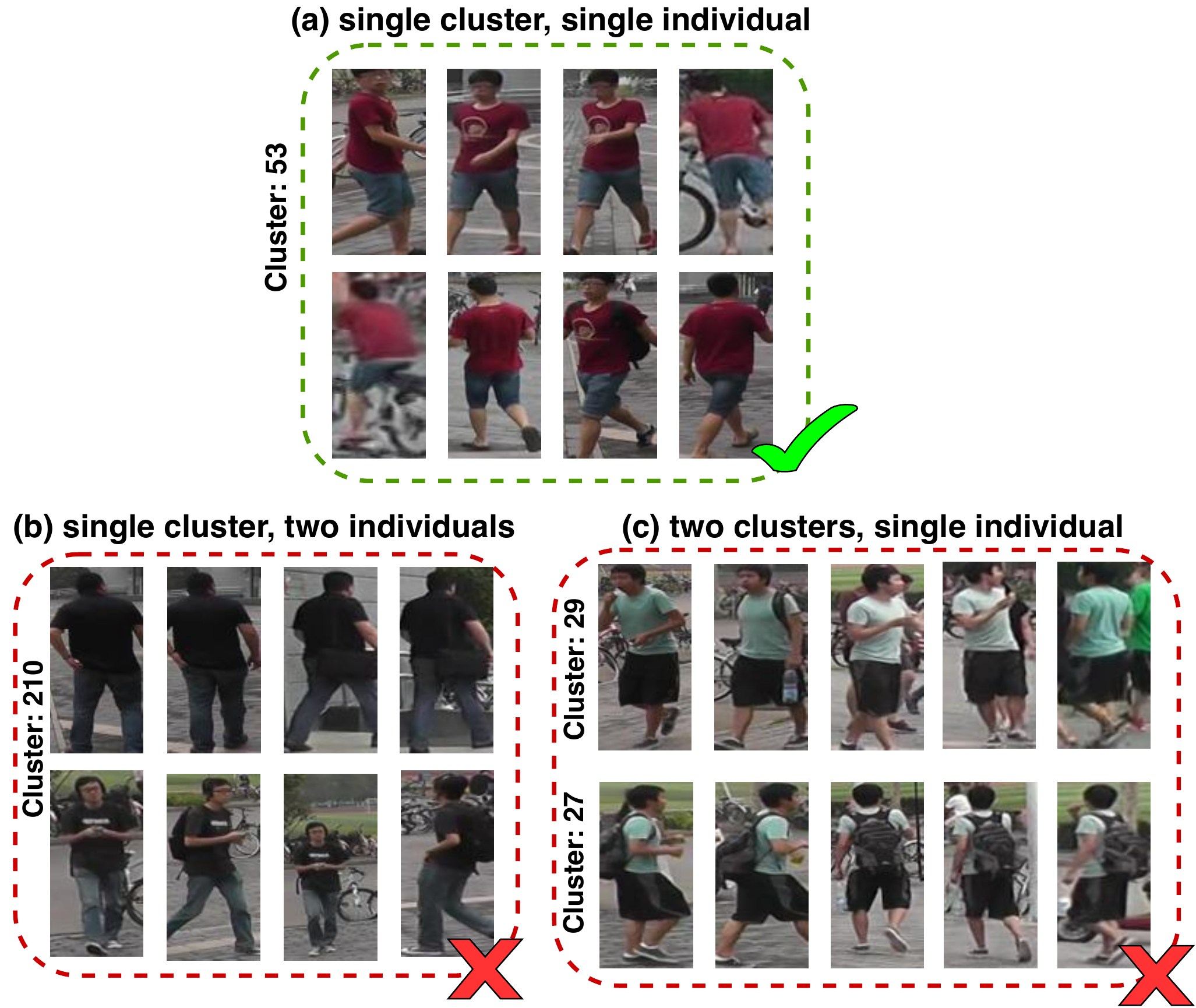}
\caption{Different types of clusters arising from the proposed ktCUDA algorithm - (a) Good cluster (GC): A cluster containing single individual who does not appear in any other clusters, (b) Mixed cluster (MC): a cluster with two or more different individuals and (c) Divided clusters (DC): an individual is split across two or more different clusters. Best viewed in color.
}
\label{fig:cluster_error}
\end{figure}

To further evaluate ktCUDA, we take a closer look at our k-reciprocal tracklet clustering. We note that k-reciprocal clustering results in three main types of clusters: Good Clusters (GC) containing only a single individual who does not appear in any other clusters, mixed clusters (MC) where multiple different individuals are in a single cluster and divided clusters (DC) where a single individual appears in multiple clusters. An example of the three types of clusters are shown in Fig.~\ref{fig:cluster_error} (a)-(c). (Note there is also a third error type which is a mix of MC and DC.)

\begin{figure}[t]
\centering
\includegraphics[width=0.48\textwidth]{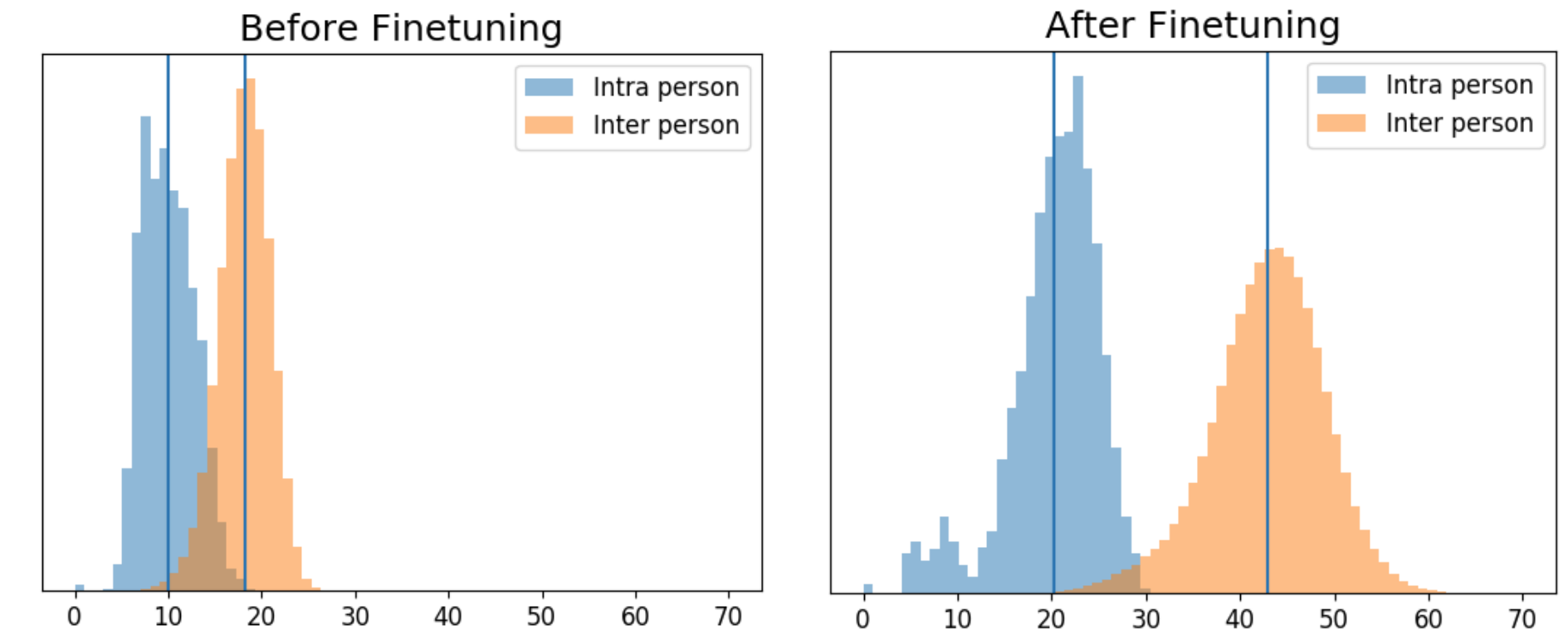}

\caption{Separation of clusters with same vs different individuals. Distance between clusters with same individuals (divided clusters Fig.~\ref{fig:cluster_error}) shown as \textit{Intra person} and distance between clusters with different individuals shown as \textit{Inter person}. Plots shown before triplet loss fine-tuning (left) and after fine-tuning (right). During fine-tuning it can be seen that the \textit{Inter person} clusters are pushed further away than \textit{Intra person} clusters.}
\label{fig:cluster_dist}
\end{figure}

Interestingly, of the two types of errors --mixed clusters and divided clusters-- we find that the presence of divided clusters doesn't negatively impact triplet loss fine-tuning. If we plot the distance between divided clusters (a.k.a. intra person) and distance between clusters with different individuals (a.k.a. inter person) before and after fine-tuning (Fig.~\ref{fig:cluster_dist}), we see both distances increase but the inter person distances increases more than intra person distance. Meaning even with the presence of divided clusters, the triplet loss is able to separate different individuals because triplet loss is not directly forcing different individuals closer. However, mixed clusters do present a problem as that will force different individuals closer.


\section{Conclusion}
In this work, we presented new strategies for unsupervised person re-ID using unlabelled data
from a target domain. Our method addressed the two main limitations of the current domain adaptation approaches: first, using source domain distance metrics for pseudo-labelling in target domain and second, relying heavily on limited source domain data. The two problems were addressed by the proposed k-reciprocal tracklet Clustering for Unsupervised Domain Adaptation (ktCUDA) method and the proposed comprehensive Synthesized Heterogeneous RE-id Domain (SHRED), respectively. Addressing these issues allowed the presented ktCUDA method to become more scalable for real-world applications. Extensive evaluation was done on image and video person re-ID benchmark datasets to validate the effectiveness of the proposed ktCUDA in outperforming other state-of-the art unsupervised domain adaptation methods in person re-ID.


{\small
\bibliographystyle{ieee}
\bibliography{egbib}
}

\end{document}